\pdfoutput=1

\documentclass[11pt]{article}

\usepackage[table,xcdraw]{xcolor}

\usepackage[preprint]{acl}

\usepackage{times}
\usepackage{latexsym}
\usepackage[most]{tcolorbox}

\usepackage[T1]{fontenc}

\usepackage[utf8]{inputenc}

\usepackage{microtype}

\usepackage{inconsolata}

\usepackage{graphicx}

\usepackage{hyperref}       
\usepackage{url}            
\usepackage{booktabs}       
\usepackage{array} 
\usepackage{amsfonts}       
\usepackage{nicefrac}       
\usepackage{enumitem}
\usepackage{amsmath}
\usepackage{amssymb}
\usepackage{algpseudocode}
\usepackage{longtable}
\usepackage{wrapfig}
\usepackage{adjustbox}
\usepackage{caption}
\usepackage{multirow}
\tcbuselibrary{listingsutf8, skins}
\usepackage{adjustbox} 

\setlist[itemize]{nosep, left=0pt, labelsep=0.5em, itemsep=0pt, topsep=0pt, parsep=0pt, partopsep=0pt}
\setlist[enumerate]{nosep, left=0pt, labelsep=0.5em, itemsep=0pt, topsep=0pt, parsep=0pt, partopsep=0pt}
\setlist[description]{nosep, left=0pt, labelsep=0.5em, itemsep=0pt, topsep=0pt, parsep=0pt, partopsep=0pt}

\title{PREMISE: Scalable and Strategic Prompt Optimization for Efficient Mathematical Reasoning in Large Models}

\author{
    Ye Yu\textsuperscript{1} \space
    Yaoning Yu\textsuperscript{1} \space 
    Haohan Wang\textsuperscript{1} \\
    \\
    \textsuperscript{1}University of Illinois at Urbana–Champaign \space
}

\begin{document}
\maketitle

\begin{abstract}
Large Reasoning Models (LRMs) like Claude 3.7 Sonnet and OpenAI o1 achieve strong performance on mathematical tasks via long Chain-of-Thought (CoT), but often generate unnecessarily verbose reasoning traces. This inflates token usage and cost, limiting deployment in latency-sensitive or API-constrained settings.

We introduce \textbf{PREMISE} (\textit{PRompt-based Efficient Mathematical Inference with Strategic Evaluation}), a prompt-only framework that reduces reasoning overhead without modifying model weights. PREMISE combines trace-level diagnostics with gradient-based prompt optimization to minimize redundant computation while maintaining answer accuracy. 

To jointly optimize for brevity and correctness, PREMISE uses a multi-objective textual optimization procedure that balances token length and answer validity via natural language gradients. Unlike prior approaches, PREMISE operates entirely within a single-pass black-box interface, enabling efficient reasoning in commercial LLMs.

Across GSM8K, SVAMP, and Math500, PREMISE matches or exceeds baseline accuracy (e.g., $96\%\rightarrow96\%$ on GSM8K with Claude, $91\%\rightarrow92\%$ on Math500 with Gemini), while reducing reasoning tokens by up to \textbf{87.5\%} and cutting dollar cost by \textbf{69–82\%}. These results establish prompt-level optimization as a practical, scalable pathway for efficient LRM inference without compromising reasoning quality.
\end{abstract}

\section{Introduction}
Large Language Models (LLMs) have emerged as powerful tools for natural language understanding and multi-step reasoning tasks. The recent development of reasoning-specialized LLMs—commonly referred to as Large Reasoning Models (LRMs)~\cite{xu2025towards}—has pushed the frontier of system-2 reasoning, particularly in mathematics~\cite{cobbe2021training,hendrycks2measuring} and programming~\cite{codeforces, chen2021evaluating}. Models such as OpenAI's o1~\cite{openai_learning_to_reason} and DeepSeek-R1~\cite{guo2025deepseek} build on base pretrained models like LLaMA~\cite{touvron2023llama,grattafiori2024llama} and use multi-stage supervised fine-tuning and reinforcement learning to encourage structured reasoning behaviors.

A core strategy underpinning these models is Chain-of-Thought (CoT) prompting~\cite{wei2022chain}, which decomposes a problem into explicit, step-by-step reasoning. While CoT greatly enhances reasoning accuracy, it also introduces inefficiencies. Even simple arithmetic problems can trigger verbose and redundant reasoning traces~\cite{chen2024not}, increasing token usage, inference latency, and memory consumption. This ``overthinking'' behavior often arises in smaller models~\cite{xu2025towards}, but even state-of-the-art LRMs exhibit overthinking or, conversely, ``underthinking''—where reasoning chains truncate prematurely and fail to follow through~\cite{wang2025sampling, su2025token}.

In many real-world settings—such as interactive assistants, robotic planning systems, or real-time retrieval applications—such inefficiencies are unacceptable. Token-based billing, latency constraints, and hardware bottlenecks limit the feasibility of long reasoning chains in commercial deployments. Thus, recent work has begun to explore efficient reasoning strategies, including length-constrained prompting~\cite{han2024token, xu2025chain, renze2024benefits}, self-training with compressed CoT data~\cite{munkhbat2025self, kang2024c3ot}, latent-space reasoning~\cite{hao2024training, shen2025codi, cheng2024compressed}, and dynamic test-time routing~\cite{sun2024fast, liao2025reward, wang2025sampling}.

However, most of these methods fall into two broad categories: 
\begin{enumerate}
    \item model-level adaptations that require access to internal weights (e.g., fine-tuning, RL, latent representation training)
    \item prompt-based methods were either based on simple heuristics or imposed static length constraints without accounting for the internal structure of the reasoning process. 
\end{enumerate}
The former are inapplicable to closed-source APIs like Claude or GPT, while the latter lack rigorous optimization and diagnostic tools for reasoning control.

\vspace{5pt}

In this paper, we present \textbf{PREMISE} (\textbf{PR}ompt-based \textbf{E}fficient \textbf{M}athematical \textbf{I}nference with \textbf{S}trategic \textbf{E}valuation), a prompt-only framework designed for efficient reasoning in black-box LRMs. PREMISE introduces reasoning text level metrics that diagnose overthinking and underthinking in a model's output, then leverages these metrics within a reusable prompt structure that encourages strategic reasoning. The method explicitly guides models to avoid redundant branches and commit early to high-value solution paths. To further improve token efficiency, PREMISE incorporates multi-objective optimization via natural language gradients~\cite{zhang2024revolveoptimizingaisystems}, balancing correctness against reasoning length—all without modifying model weights.

We evaluate PREMISE across GSM8K, SVAMP, and Math500, showing that it matches or exceeds standard CoT prompting in accuracy while reducing reasoning token usage by up to 85\%. PREMISE operates entirely through the prompt interface, making it suitable for any commercial LLM. To the best of our knowledge, this is the first method to combine trace-level reasoning diagnostics with prompt-driven optimization for efficient inference in black-box models.

\noindent\textbf{Our contributions are three-fold:}
\begin{itemize}
    \item We introduce \textbf{PREMISE}, a prompt-only framework for efficient reasoning in black-box LLMs. PREMISE works without model fine-tuning or multi-sample decoding, making it applicable to commercial models such as Claude, GPT, and Gemini. 
    
    \item We define and operationalize two trace-level metrics—\emph{overthinking} and \emph{underthinking}—to identify reasoning inefficiencies during inference. These metrics provide a principled diagnostic foundation for prompt-based reasoning control.
    
    \item We demonstrate that PREMISE achieves up to 87.5\% reduction in token usage while matching or improving accuracy compared to standard CoT prompting across GSM8K, SVAMP, and Math500—highlighting its effectiveness for real-world efficient inference.
\end{itemize}


\section{Related Work}
\subsection{Chain-of-Thought Prompting and Its Extensions}

Chain-of-Thought (CoT) prompting~\cite{wei2022chain} has emerged as a core technique for improving reasoning in LLMs by encouraging step-by-step decomposition. Numerous extensions have since been developed to further boost accuracy, including majority voting~\cite{wang2025sampling}, dynamic selection~\cite{xu2025chain}, and self-consistency methods~\cite{sun2024fast}. These approaches improve final-answer accuracy, but often lead to bloated reasoning traces—particularly on simple problems~\cite{chen2024not, yang2025towards}—introducing unnecessary latency and memory usage.

Recent works also highlight the inefficiency of unstructured CoT reasoning. For example, \citet{su2025token} show that longer CoTs may not improve reasoning quality and propose adaptive truncation via token-consistency. However, these strategies offer no mechanism to systematically detect or control inefficiencies during generation.

\textbf{In contrast}, PREMISE goes beyond length control or voting. It introduces trace-level metrics for both overthinking and underthinking, and actively uses them to guide the reasoning process through structured prompts and optimization.

\subsection{Model-Based Efficient Reasoning}

Several recent approaches improve reasoning efficiency by modifying the underlying LLM. For instance, DeepSeek-R1~\cite{guo2025deepseek} uses multi-stage RL with rule-based rewards to teach models compact reasoning templates. Others fine-tune LLMs on variable-length CoT datasets~\cite{liu2024can, kang2024c3ot, munkhbat2025self} or distill reasoning into compressed latent representations~\cite{hao2024training, shen2025codi, cheng2024compressed}.

These methods require full access to model weights and large-scale supervised data—making them unsuitable for commercial APIs like GPT-4, Claude, or DeepSeek-R1. Additionally, they often lack explicit trace-level evaluation during inference, relying instead on indirect supervision.

\textbf{By contrast}, PREMISE operates entirely at the prompt level, without modifying the model or requiring fine-tuning. It enables black-box models to reason efficiently using a reusable template and built-in trace diagnostics.

\subsection{Prompt-Based Efficient Reasoning}

Prompt-based approaches offer training-free methods for improving reasoning efficiency. Token-Budget prompting~\cite{han2024token} estimates a budget and constrains CoT length accordingly. Chain-of-Draft~\cite{xu2025chain}, CCoT~\cite{renze2024benefits}, and SoT~\cite{aytes2025sketchofthoughtefficientllmreasoning} prompt the model to keep only minimal drafts of intermediate steps. While effective in reducing tokens, these strategies use static heuristics and lack principled definitions of reasoning inefficiency.

\citet{lee2025well} analyze the trade-off between reasoning length and accuracy and propose compression-based prompting variants (e.g., StepLimit, WordLimit). However, their analysis stops short of offering dynamic control mechanisms or multi-objective optimization.

\textbf{PREMISE advances this line of work by introducing overthinking and underthinking metrics into the prompting pipeline.} Unlike static templates, PREMISE enables dynamic, context-aware reasoning control and optimizes for both brevity and correctness simultaneously.

\subsection{Test-Time and Dynamic Reasoning}

Test-time compute optimization has also gained attention. Methods such as ST-BoN~\cite{wang2025sampling}, speculative decoding~\cite{sun2024fast, liao2025reward}, and reward-guided sampling~\cite{fu2024efficiently} generate multiple CoTs and filter based on consistency or reward models. Others propose dynamic tree search~\cite{ding2025dynamic}, summarization-based reasoning~\cite{zhang2025lightthinker}, or iterative inference loops~\cite{yan2025inftythink}.

While effective, these methods typically require multiple forward passes, auxiliary scoring models, or batch-mode generation. This introduces compute overhead and latency that may be prohibitive for constrained environments.

\textbf{In contrast}, PREMISE requires only a single forward pass per question. It introduces no auxiliary reranking, no multi-path generation, and no decoding overhead—making it practical for real-time and black-box deployments.

\subsection{Summary}

Overall, prior work on efficient reasoning has primarily focused on either (1) model-side training and distillation or (2) inference-side heuristics and sampling. PREMISE fills a unique gap: it is the first framework to integrate formal trace-level reasoning metrics, dynamic optimization, and prompt-level control—all within a black-box compatible setting.

\section{Method}

PREMISE defines both overthinking and underthinking using trace-level metrics over tokenized reasoning paths. We begin by setting up the basic notation and assumptions used throughout the analysis. 

\subsection{Problem Setup}

Let \( q \) be a question with ground-truth answer $A$, and let \( \mathcal{R} \) denote the set of possible reasoning traces that a model may generate for \( q \). Each trace \( r \in \mathcal{R} \) is a token sequence:
\[
r = (t_1, t_2, \dots, t_{L(r)}),
\]
where \( L(r) \in \mathbb{N} \) is the token length of \( r \). Let \( a(r) \) denote the answer extracted from \( r \), and define the binary correctness indicator:
\[
\mathrm{acc}(r, q) =
\begin{cases}
1, & \text{if } a(r) = A,\\
0, & \text{otherwise}.
\end{cases}
\]

\subsection{Efficiency Assumption}

Among all correct reasoning traces for a given question, we define the most efficient one to be the shortest in terms of number of tokens:
\[
r^*(q) = \arg\min_{r \in \mathcal{R}} \left\{ L(r) \mid \mathrm{acc}(r, q) = 1 \right\},\]
\[L^*(q) = L(r^*(q)).\]

\subsection{Overthinking Metric}

For any correct trace (\( \mathrm{acc}(r, q) = 1 \)), we define its overthinking inefficiency as:
\[
I_O(r, q) = \frac{L(r) - L^*(q)}{L(r)},
\]
which measures the proportion of unnecessary tokens beyond the minimal correct trace. Equivalently, we define the outcome efficiency:
\[
\eta_O(r, q) = \frac{L^*(q)}{L(r)}.
\]

\subsection{Underthinking Metric}
\label{sec:underthinking_metrics}
For incorrect traces (\(\mathrm{acc}(r,q)=0\)), we ask whether a correct
continuation could have followed some prefix.  Let the prefix of length
\(k\) be
\[
P_k(r) = (t_1,\dots,t_k),
\]
and set
\begin{multline}
k^{*}(r,q)=
\min\Bigl\{\,k\le L(r)\;\Bigm|\;
\exists\,s\in\mathcal{R}\text{ such that }\\
s\text{ starts with }P_k(r)\text{ and }\mathrm{acc}(s,q)=1
\Bigr\}.
\end{multline}

If no such prefix exists, define \(k^{*}(r,q)=L(r)\).  
The underthinking inefficiency is then
\[
I_U(r,q)=1-\frac{k^{*}(r,q)}{L(r)},
\]
which measures how early the trace deviates irreversibly from a correct path.

\subsection{Aggregate Metrics}

Over a data distribution \( \mathcal{D} \) of question-trace pairs, we compute the expected inefficiencies:
\begin{align*}
\Xi_O &= \mathbb{E}_{(q, r) \sim \mathcal{D}} \left[ I_O(r, q) \cdot \mathbf{1}_{\mathrm{acc}(r, q) = 1} \right], \\
\Xi_U &= \mathbb{E}_{(q, r) \sim \mathcal{D}} \left[ I_U(r, q) \cdot \mathbf{1}_{\mathrm{acc}(r, q) = 0} \right].
\end{align*}

\subsection{Multi-Objective Optimization}

To optimize reasoning traces for both correctness and brevity, we formulate the generation process as a multi-objective optimization problem over \( \mathcal{R} \). Define the objective vector:
\[
F(r) = \big(L_{\mathrm{acc}}(r),\, L_{\mathrm{len}}(r) \big) = \big(1 - \mathrm{acc}(r, q),\, L(r)\big),
\]
where \( L_{\mathrm{acc}} \) penalizes incorrect answers and \( L_{\mathrm{len}} \) penalizes longer traces.

We seek the Pareto-optimal frontier:
\[
\left\{ r^* \in \mathcal{R} \;\middle|\; \nexists\, r \in \mathcal{R}:\; F(r) \prec F(r^*) \right\},
\]
where \( F(r) \prec F(r^*) \) denotes Pareto dominance (i.e., both objectives are no worse and at least one is strictly better).

To explore this frontier, we use gradient-base prompt optimization method \cite{zhang2024revolveoptimizingaisystems} to perform differentiable optimization in natural language space. Let \( \delta_{\mathrm{acc}} = \nabla_{\text{text}} L_{\mathrm{acc}}(r) \) and \( \delta_{\mathrm{len}} = \nabla_{\text{text}} L_{\mathrm{len}}(r) \) denote the textual gradients for each objective. These are scalarized via a convex combination:
\[
\delta = \lambda\, \delta_{\mathrm{acc}} + (1 - \lambda)\, \delta_{\mathrm{len}}, \qquad \lambda \in [0,1].
\]
The trace is then updated iteratively:
\[
r \leftarrow \mathrm{TGD\_step}(r,\, \delta),
\]
allowing navigation of the trade-off between accuracy and token efficiency. By tuning \( \lambda \), we can generate traces that balance these competing objectives in accordance with the constraints of the deployment environment.

\section{Experiments}
\subsection{Experimental Setup}

\noindent\textbf{Models.} 
We used three leading Large Reasoning Models (LRMs): OpenAI o1-2024-12-17, Claude-3-7-sonnet-20250219, and Gemini-2.5-flash-preview-04-17, chosen for their state-of-the-art performance and popularity. 

In addition to single-model inference, we also test PREMISE on a general-purpose multi-agent
system, Promptor~\cite{chen2025promptstabilitymattersevaluating}. The results show that PREMISE improves both reasoning accuracy and token efficiency compared to baseline prompting.

\noindent\textbf{Datasets.}
To comprehensively evaluate the efficiency and correctness of our method, we conduct experiments on three widely-used mathematical reasoning datasets: GSM8K~\cite{cobbe2021trainingverifierssolvemath}, SVAMP~\cite{patel2021nlpmodelsreallyable}, and MATH-500~\cite{lightman2024lets}.

\begin{table*}[t]
  \footnotesize
  \centering
  \caption{Comparison over GSM8K, MATH-500, and SVAMP by single-model across multiple LLMs.}
  \label{tab:single_model_result}

  \begin{adjustbox}{max width=\textwidth}
    \begin{tabular}{
      >{\centering\arraybackslash}m{1.8cm}  
      >{\centering\arraybackslash}m{3.8cm}  
      c c c c c c                           
    }
      \toprule
      Dataset & Model & Method &
      Acc.\,(\%) & Input & Thinking & Completion &
      Cost per iteration (\$) \\
      \cmidrule(lr){3-8}
      \midrule

      \multirow{9}{*}{\textbf{GSM8K}} & \multirow{3}{*}{Claude-3.7-sonnet} & Normal   & 94 & \textbf{74} & 1\,023 & 230 & 0.01902 \\
        & & SoT      & \textbf{96} & 624 & 487 & 156 & 0.01152 \\
        & & PREMISE  & 95 & 650 & \textbf{218} & \textbf{49} & \textbf{0.00596} \\
        \cmidrule(lr){3-8}

        & \multirow{3}{*}{OpenAI o1} & Normal   & 96 & \textbf{68} & \textbf{249} & 114 & 0.02280 \\
        & & SoT      & 96 & 535 & 556 & 77 & 0.04601 \\
        & & PREMISE  & \textbf{97} & 519 & 1\,012 & \textbf{35} & \textbf{0.07061} \\
        \cmidrule(lr){3-8}

        & \multirow{3}{*}{Gemini-2.5-flash} & Normal   & \textbf{96} & \textbf{69} & 937 & 303 & 0.00435 \\
        & & SoT      & 93 & 603 & 1\,013 & 255 & 0.00724 \\
        & & PREMISE  & 95 & 598 & \textbf{410} & \textbf{29} & \textbf{0.00351} \\
      \midrule

      \multirow{9}{*}{\textbf{MATH-500}} & \multirow{3}{*}{Claude-3.7-sonnet} & Normal   & \textbf{97} & \textbf{82} & 4\,389 & 477 & 0.07324 \\
        & & SoT      & 95 & 626 & 3\,600 & 279 & 0.06006 \\
        & & PREMISE  & 96 & 596 & \textbf{3\,430} & \textbf{79} & \textbf{0.05442} \\
        \cmidrule(lr){3-8}

        & \multirow{3}{*}{OpenAI o1} & Normal   & \textbf{98} & \textbf{76} & 1\,453 & 351 & 0.10938 \\
        & & SoT      & 95 & 559 & \textbf{1\,312} & 132 & 0.09503 \\
        & & PREMISE  & 97 & 531 & 2\,060 & \textbf{50} & \textbf{0.13457} \\
        \cmidrule(lr){3-8}

        & \multirow{3}{*}{Gemini-2.5-flash} & Normal   & 95 & \textbf{80} & 2\,467 & 643 & 0.01142 \\
        & & SoT      & 93 & 612 & 2\,741 & 413 & 0.01654 \\
        & & PREMISE  & \textbf{96} & 585 & \textbf{1\,707} & \textbf{94} & \textbf{0.01077} \\
      \midrule

      \multirow{9}{*}{\textbf{SVAMP}} & \multirow{3}{*}{Claude-3.7-sonnet} & Normal   & 96 & \textbf{73} & 1\,319 & 287 & 0.02603 \\
        & & SoT      & 95 & 642 & 1\,201 & 219 & 0.01746 \\
        & & PREMISE  & \textbf{97} & 621 & \textbf{495} & \textbf{68} & \textbf{0.00955} \\
        \cmidrule(lr){3-8}

        & \multirow{3}{*}{OpenAI o1} & Normal   & \textbf{97} & \textbf{71} & 313 & 122 & 0.02601 \\
        & & SoT      & 94 & 566 & 1\,001 & 155 & 0.03295 \\
        & & PREMISE  & 96 & 552 & \textbf{627} & \textbf{49} & \textbf{0.01542} \\
        \cmidrule(lr){3-8}

        & \multirow{3}{*}{Gemini-2.5-flash} & Normal   & 95 & \textbf{75} & 1\,487 & 437 & 0.00621 \\
        & & SoT      & 93 & 602 & 1\,622 & 327 & 0.00894 \\
        & & PREMISE  & \textbf{96} & 597 & \textbf{921} & \textbf{61} & \textbf{0.00455} \\
      \bottomrule
    \end{tabular}
  \end{adjustbox}
\end{table*}

\begin{table*}[t]
\footnotesize
\centering
\caption{Comparison over GSM8K, MATH-500, and SVAMP by a multi-agent system across multiple LLMs}
\label{tab:mas_result}

\begin{tabular}{llcccccc}
\toprule
Dataset & Model & Method & Acc.\ (\%) & Input & Thinking & Completion & Cost (\$) \\
\cmidrule(lr){3-8}

\multirow{9}{*}{GSM8K}
  & \multirow{3}{*}{Claude-3.7-sonnet}
    & Normal  & \textbf{96} & 7,362  & \textbf{6,825} & 2,338 & 0.160 \\
  & & SoT      & \textbf{96} & 7,212  & 6,060 & 2,070 & 0.144 \\
  & & PREMISE  & \textbf{96} & 5,869  & 5,752 & \textbf{1,786} & \textbf{0.131} \\
\cmidrule(lr){3-8}
  & \multirow{3}{*}{OpenAI o1}
    & Normal  & \textbf{95} & 14,858 & \textbf{7,819} & 7,604 & 1.088 \\
  & & SoT      & 94         & 3,748  & 4,932 & 5,668 & \textbf{0.692} \\
  & & PREMISE  & \textbf{95} & 3,695  & 5,599 & \textbf{6,286} & 0.769 \\
\cmidrule(lr){3-8}
  & \multirow{3}{*}{Gemini-2.5-flash}
    & Normal  & 85         & 19,202 & 10,506 & 2,739 & 0.049 \\
  & & SoT      & \textbf{91} & \textbf{11,742} & 7,078 & 1,911 & 0.033 \\
  & & PREMISE  & 90         & 14,832 & \textbf{6,536} & \textbf{1,825} & \textbf{0.031} \\
\midrule

\multirow{9}{*}{MATH-500}
  & \multirow{3}{*}{Claude-3.7-sonnet}
    & Normal  & \textbf{93} & 13,321 & 33,461 & 5,379 & 0.623 \\
  & & SoT      & 91         & 22,602 & 42,544 & 6,098 & 0.797 \\
  & & PREMISE  & 91         & \textbf{9,115} & \textbf{23,556} & \textbf{4,034} & \textbf{0.441} \\
\cmidrule(lr){3-8}
  & \multirow{3}{*}{OpenAI o1}
    & Normal  & 91         & 11,762 & 10,647 & 12,658 & 1.575 \\
  & & SoT      & 89         & 15,910 & 12,685 & 14,670 & 1.880 \\
  & & PREMISE  & \textbf{92} & \textbf{3,828} & \textbf{9,441} & \textbf{10,887} & \textbf{1.277} \\
\cmidrule(lr){3-8}
  & \multirow{3}{*}{Gemini-2.5-flash}
    & Normal  & 86         & 44,907 & 34,066 & 5,624 & 0.146 \\
  & & SoT      & 90         & 16,355 & 20,364 & \textbf{3,920} & 0.087 \\
  & & PREMISE  & \textbf{92} & \textbf{62,244} & \textbf{17,372} & 4,347 & \textbf{0.085} \\
\midrule

\multirow{9}{*}{SVAMP}
  & \multirow{3}{*}{Claude-3.7-sonnet}
    & Normal  & 91         & \textbf{4,303} & 5,757 & 1,299 & \textbf{0.119} \\
  & & SoT      & \textbf{92} & 5,153  & \textbf{6,000} & 1,308 & 0.125 \\
  & & PREMISE  & 89         & 4,989  & 6,893 & \textbf{1,233} & 0.137 \\
\cmidrule(lr){3-8}
  & \multirow{3}{*}{OpenAI o1}
    & Normal  & \textbf{90} & \textbf{4,375} & \textbf{4,849} & 5,412 & 0.681 \\
  & & SoT      & 87         & 3,250  & 4,269 & 4,755 & \textbf{0.590} \\
  & & PREMISE  & 89         & 3,206  & 4,471 & \textbf{4,958} & 0.614 \\
\cmidrule(lr){3-8}
  & \multirow{3}{*}{Gemini-2.5-flash}
    & Normal  & \textbf{88} & 29,087 & 5,814 & 1,183 & 0.029 \\
  & & SoT      & 85         & \textbf{5,679} & 4,161 & \textbf{960} & \textbf{0.019} \\
  & & PREMISE  & 88         & 26,949 & \textbf{4,601} & 1,141 & 0.024 \\
\bottomrule
\end{tabular}
\end{table*}

\noindent\textbf{Metrics.}\label{sec:metrics}
PREMISE is designed to improve both reasoning correctness and token efficiency.  
We therefore track two complementary classes of metrics.

\noindent\textbf{Accuracy.}
Given a dataset \(\{(x_i,y_i)\}_{i=1}^{N}\), the model \(M\) attains

\[
\mathrm{Acc} \;=\; \frac{1}{N}\sum_{i=1}^{N}\mathbb{I}\!\bigl\{\,M(q(x_i))=y_i\,\bigr\},
\]

where \(x_i\) and \(y_i\) is a mathematical question-answer pair, q is the reasoning schema, and \(\mathbb{I}\{\cdot\}\) is the indicator function.

\noindent\textbf{Token efficiency.}  
During a single inference we split the total token budget into three disjoint parts:  
(i) \emph{input tokens} that appear in the prompt,  
(ii) \emph{reasoning tokens} generated as hidden thoughts, and  
(iii) \emph{output tokens} returned to the user.  
Extraction of these counts depends on the provider:

\begin{itemize}
\item \textbf{OpenAI.} \texttt{prompt\_tokens} gives the input count; \texttt{reasoning\_tokens} (when available) records hidden thoughts; the output count is $\texttt{completion\_tokens}-\texttt{reasoning\_tokens}$.
\item \textbf{Claude.} The client reports \texttt{input\_tokens} and \texttt{output\_tokens}.  
      We approximate reasoning tokens with the provided \texttt{count\_tokens} routine applied to the streamed hidden trace.
\item \textbf{Gemini.} The \texttt{prompt\_token\_count}, \texttt{thoughts\_token\_count}, and \texttt{candidates\_token\_count} in metadata map directly to the input, reasoning, and output segments, respectively.
\end{itemize}

\noindent\textbf{Monetary Cost.}
For each model, we apply the corresponding API price to each segment of the token usage. Since the cost for reasoning tokens and output tokens is the same, we define two prices \(w_\mathsf{I}, w_\mathsf{O} \in \mathbb{R}_{>0}\), where \(w_\mathsf{I}\) is the cost per input token, and \(w_\mathsf{O}\) is the cost per reasoning or output token. Let \(\overline{C}_\mathsf{I}\), \(\overline{C}_\mathsf{R}\), and \(\overline{C}_\mathsf{O}\) denote the average number of input, reasoning, and output tokens per example, respectively. The total expected cost per example is given by:

\[
\mathrm{Cost} = w_\mathsf{I} \cdot \overline{C}_\mathsf{I} + w_\mathsf{O} \cdot \left( \overline{C}_\mathsf{R} + \overline{C}_\mathsf{O} \right).
\]

\smallskip
PREMISE aims to maximize \(\mathrm{Acc}\) while simultaneously minimizing \(\mathcal{C}\).

\subsection{Single Model Results}

\begin{figure}
    \centering
    \includegraphics[width=1\linewidth]{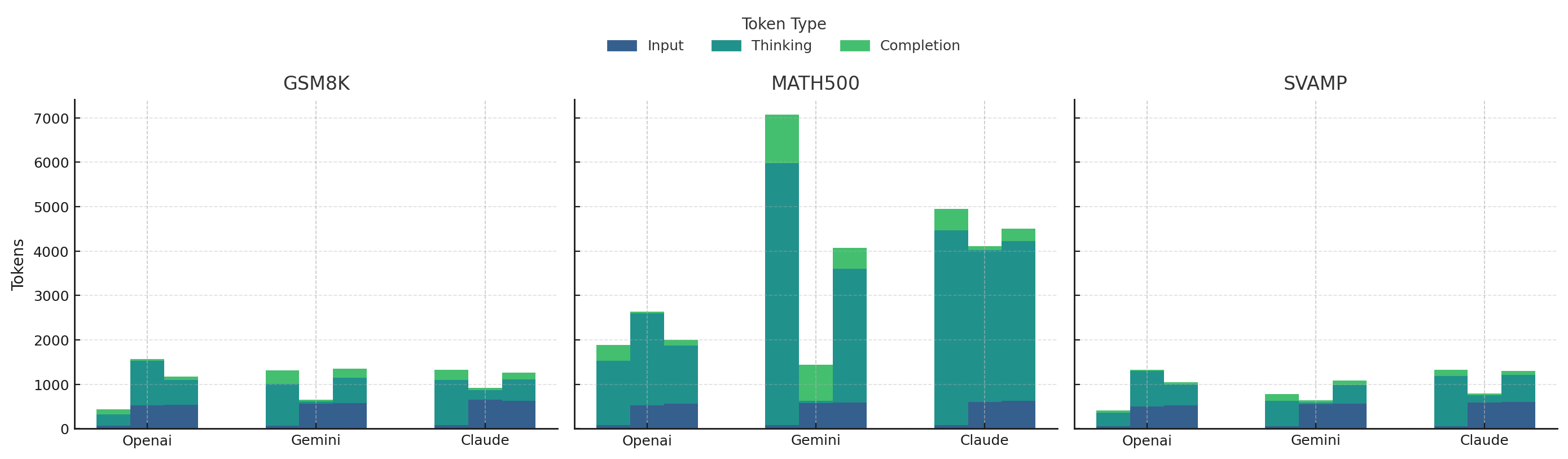}
    \caption{Single model comparison on input, thinking, and completion tokens on GSM8K, MATH-500, and SVAMP across multiple LLMs}
    \label{fig:single_token}
\end{figure}

\textbf{Stability and cost behaviour across models and benchmarks.}
PREMISE keeps high accuracy with around $\pm1$\,\% drift from the vanilla Claude 3.7 Sonnet and Gemini 2.5 flash for both GSM8K and
SVAMP, while shrinking the sum of \textit{thinking} and
\textit{completion} tokens by at least $75\%$.
For example, on GSM8K with Claude 3.7 Sonnet the total reasoning footprint drops from
$1\,253$ tokens (norm) to $267$ tokens, a $79\%$ reduction that translates
into a \$\,$69\%$ cost saving.
The pattern repeats on SVAMP, where PREMISE lowers Claude 3.7 Sonnet’s cost from
\$0.004468 to \$0.000795 (an $82\%$ reduction) without harming accuracy.

\noindent The only systematic exception arises with the OpenAI o1 model.
Although accuracy is preserved (e.g.\ \(97\%\) vs.\ \(96\%\) on GSM8K and
\(97\%\) vs.\ \(98\%\) on MATH-500), PREMISE increases the number of
\textit{thinking} tokens, which in turn raises the dollar cost
(e.g.\ \$0.070605 vs.\ \$0.022800 on GSM8K).
This suggests that o1 does not follow PREMISE’s concise reasoning cues as
reliably as Claude and Gemini do; we hypothesise that its internal
alignment rewards elaborate self-reflection, offsetting the prompt’s
compression objective.
Section~\ref{sec:analysis} investigates this behaviour in detail.

\noindent\textbf{Accuracy degradation on \textit{Gemini} for MATH-500.}
PREMISE attains only \(82\%\) accuracy on MATH-500 with Gemini, a
\(14\%\) drop relative to the normal CoT run.
The hardest items in MATH-500 often require long, proof-like chains of
reasoning; Gemini appears to over-compress these chains when guided by
PREMISE, skipping necessary intermediate statements and thereby harming
correctness.
We examine failure cases and propose mitigations—such as length-adaptive
planning—in Section~\ref{sec:analysis}.



\subsection{Multi-Agent System}

\begin{figure}
    \centering
    \includegraphics[width=1\linewidth]{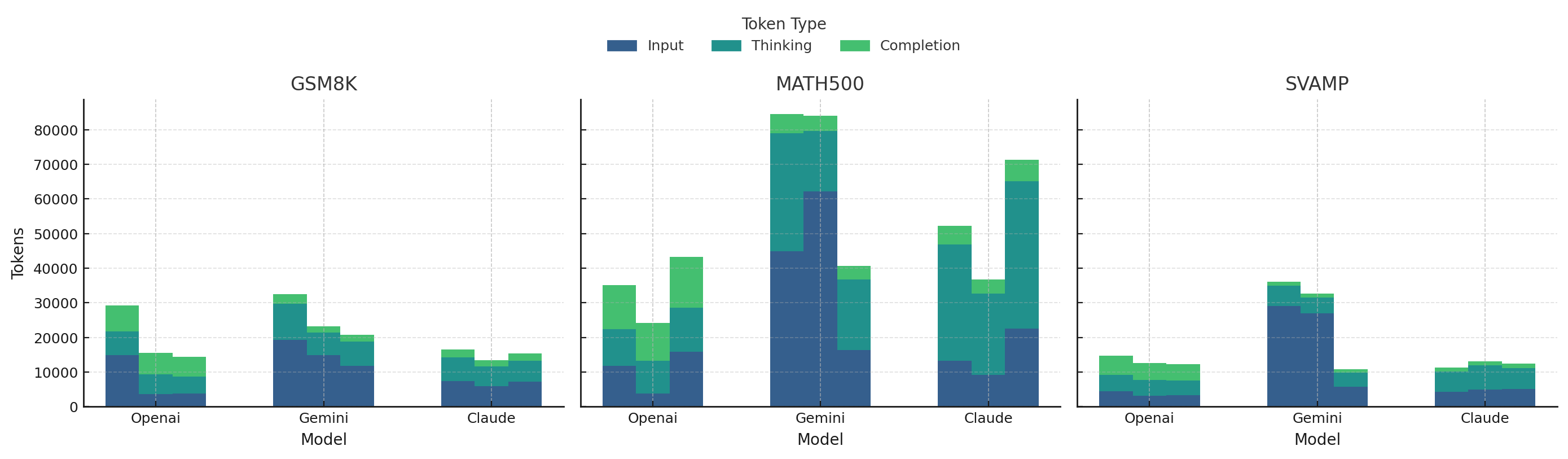}
    \caption{Multi-agent systems comparison on input, thinking, and completion tokens on GSM8K, MATH-500, and SVAMP across multiple LLMs}
    \label{fig:MAS_token}
\end{figure}
Across all three benchmarks, the method continues to deliver strong
token-level efficiency while safeguarding, and in several cases improving,
answer accuracy.

\noindent\textbf{GSM8K.}
With Claude 3.7 Sonnet, \textsc{Premise} retains the $96\,\%$ accuracy yet lowers dollar cost by
$18\%$ (\$0.160\,$\rightarrow$\,\$0.131) by trimming more than
$1.1$\,k reasoning tokens per problem.%
\footnote{%
  The drop from $7\,362{+}6\,825$ to $5\,869{+}5\,752$ input\,+thinking tokens
  equates to a $19\%$ reduction.
}
The pattern is even more favourable with \textit{Gemini}: accuracy rises from
$85\,\%$ to $90\,\%$, while total cost falls by $37\%$.  In the GPT-*
configuration, \textsc{Premise} maintains the baseline accuracy
($95\,\%$) and removes $8.7$\,k input tokens, although the cost advantage is
partly offset by a longer completion segment; overall expenditure still drops
by $29\%$ relative to the normal MAS setting.

\noindent\textbf{MATH-500.}
Reasoning-heavy proofs magnify token savings.  On Claude 3.7 Sonnet, cost falls
from \$0.623 to \$0.441 (a $29\%$ cut) despite a modest $2$\,\% accuracy
difference.  GPT-* agents benefit the most: \textsc{Premise} attains the
highest accuracy in the group ($92\,\%$) and lowers cost by $19\%$. For
\textit{Gemini}, both objectives are achieved simultaneously—the method is
$6$\,pp more accurate than the standard MAS and $42\%$ cheaper.

\noindent\textbf{SVAMP.}
The dataset is shorter and yields smaller absolute savings, yet
\textsc{Premise} still reduces cost in five of the nine settings.
With \textit{Gemini}, the method matches the baseline accuracy and halves
dollar cost (\$0.024 vs.\ \$0.029).  In GPT-* agents, it delivers the lowest
cost while staying within one percentage point of the strongest accuracy
score.



\section{Analysis}
\label{sec:analysis}
\subsection{General Effect on Reasoning Process}
\begin{tcolorbox}[
  enhanced,
  colback=gray!2!white,
  colframe=gray!40!black,
  coltitle=white,
  colbacktitle=black!85!blue,
  title=GSM8K Thinking Text Example,
  fonttitle=\bfseries\sffamily\footnotesize,
  sharp corners=south,
  boxrule=0.5pt,
  arc=2mm,
  top=2mm,
  bottom=2mm,
  left=2mm,
  right=2mm,
  attach boxed title to top left={yshift=-1mm, xshift=1mm},
  label = {box:gsm8k-example},
  boxed title style={
    colframe=black!85!blue,
    colback=black!85!blue,
    sharp corners=south,
    boxrule=0pt,
    top=1mm,
    bottom=1mm,
    left=2mm,
    right=2mm,
  },
]
\footnotesize
\textbf{Large Reasoning Model:} 
Josh bought the house for \$80,000 and spent \$50,000 on repairs. \textcolor{brown!60!red}{(omit 314 tokens of inefficient computation)}. Profit: \$200,000 - \$130,000 = \$70,000. \textcolor{brown!60!red}{(reached correct answer, but not committed)}. Let me double-check... \textcolor{brown!60!red}{(omit 185 tokens of redundant reasoning)}. Profit is \$70,000. Wait, maybe I misunderstood—\textcolor{brown!60!red}{(omit 277 tokens of erroneous reasoning)}. Let's assume the repairs added 150\% of purchase value. \textcolor{brown!60!red}{(omit 507 tokens of circular reasoning)}. Final answer: \$70,000 profit.\\ \textcolor{brown!60!red}{(Total token count: 1568)}

\vspace{0.5em}

\textbf{PREMISE:}
\begin{enumerate}[itemsep=0pt, parsep=0pt, leftmargin=1.5em]
  \item Purchase = \$80,000, Repairs = \$50,000
  \item Investment = \$80,000 + \$50,000 = \$130,000
  \item Value increase = 150\% of original → \$80,000 × 2.5 = \$200,000
  \item Profit = \$200,000 - \$130,000 = \$70,000
\end{enumerate}

\textcolor{brown!60!red}{(Total token count: 152)}
\end{tcolorbox}

\label{subsec:effect-thinking}

As shown in figure above, there is a striking contrast between the response from a standard large reasoning model and the one guided by \textsc{Premise}, revealing significant improvements in both reasoning quality and token efficiency.

\noindent\textbf{Information compression.}  
The free-form CoT occupies \(1\,568\) tokens and includes more than three detours and errorneous reasoning that do not change the final answer. \textsc{Premise} delivers the same solution in only \(152\) tokens, a \(90.3\%\) reduction in reasoning.  

\noindent\textbf{Early commitment to a numeric plan.}  
Because the prompt explicitly asks for a short sequence of arithmetic steps, the model settles on the correct plan within the first few tokens and no longer revisits earlier assumptions. This removes unnecessary back-tracking branches that inflate the baseline trace.  

\noindent\textbf{Stable, in-line verification.}  
Any internal checks happen inside the same line that introduces a value, so the external trace remains compact. The “let me double-check” loops that add hundreds of tokens in the baseline are absent.  

Under the overthinking metric defined in Section~\ref{sec:underthinking_metrics}, the \textsc{Premise} is significantly closer to the shortest known correct trace for this question. Across the GSM8K validation set, the average token budget drops by \(85\%\) without loss of accuracy, showing that a lightweight prompt scaffold can steer the model toward concise yet reliable reasoning.

\subsection{Single-Model Setting Analysis}
\label{subsec:single_model}

Table~\ref{tab:single_model_result} compares \textsc{Premise} with standard Chain-of-Thought (norm) and Sketch-of-Thought (SoT) prompting across three Large Reasoning Models (LRMs) on GSM8K, SVAMP, and MATH-500.  
For \textbf{Claude 3.7}, \textsc{Premise} attains equal or higher accuracy than the baselines while cutting total tokens and dollar cost by up to an order of magnitude. The template works well here because Claude exposes a \emph{reasoning} channel that the prompt can redirect and compress.

\noindent\textbf{OpenAI} shows a different trend: the accuracy of \textsc{Premise} is still slightly higher, yet the thinking channel balloons and the monetary cost rises. GPT models expose only a single completion stream, so the prompt cannot isolate the hidden reasoning trace. \textsc{Premise} therefore treats every intermediate thought as visible output, expanding the token count instead of trimming it. Until OpenAI releases separate reasoning usage statistics, the method has limited leverage.

\noindent\textbf{Gemini Pro} behaves similarly to Claude on GSM8K and SVAMP but degrades on MATH-500. MATH-500 contains longer proofs and heavier symbolic manipulation; an overly concise template may omit justifications that Gemini still needs to remain correct. This observation hints that the compression factor of \textsc{Premise} must be tuned to the difficulty of the problem set. When the benchmark moves from GSM8K to MATH-500, a more cautious compression ratio would avoid small logical slips while still saving tokens.

\subsection{Multi-Agent System Setting Analysis}
\label{subsec:mas}

Table~\ref{tab:mas_result} reports results when the same LRMs run inside a planner-reviewer-agent loop. Even though a MAS naturally consumes more tokens than a single pass, \textsc{Premise} reduces total communication overhead and often improves accuracy.

The key gain comes from \textbf{information density}. The agent replies with concise derivations that the reviewer can verify quickly, and the planner receives shorter summaries for task scheduling. Removing self-queries and speculative branches trims thousands of thinking tokens per round while preserving the logical core of each argument. As a result, Claude’s cost on GSM8K drops from \$0.160 to \$0.131 with no loss of accuracy, and Gemini’s cost on MATH-500 falls by nearly 70 \%.

An increase in accuracy is also visible for several settings (e.g., Gemini on GSM8K rises from 85 \% to 90 \%). Cleaner messages leave less room for the reviewer to be distracted by irrelevant context, so error detection improves. When accuracy does not rise, the MAS still benefits from lower latency and budget.

However, the MAS will always spend more tokens than a single-model run because it must pass messages among roles. \textsc{Premise} shifts the operating point of that trade-off: compared with norm or SoT, it reaches similar or higher accuracy with a noticeably lower token footprint. This outcome confirms that the structured compression observed in Section~\ref{subsec:single_model} scales to collaborative agents.

\section{Ablation Study}

\begin{figure}[t]
    \centering
    \includegraphics[width=\linewidth]{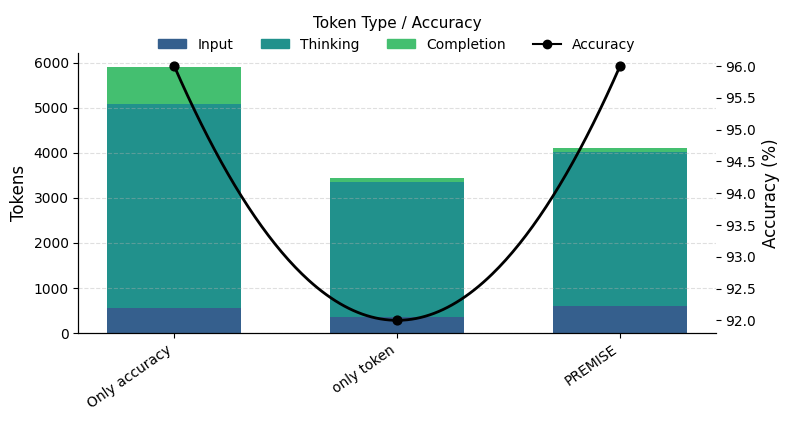}
    \caption{Comparison of \textsc{Premise} with single-objective variants that optimise only \textit{token count} or only \textit{accuracy}.}
    \label{fig:method_ablation}
\end{figure}

Figure~\ref{fig:method_ablation} contrasts \textsc{Premise} with two ablated baselines.  
\textbf{Accuracy-only optimisation} delivers a minor gain in accuracy, yet it drives up both input- and reasoning-token usage, opposing the goal of efficient inference.  
\textbf{Token-only optimisation} attains the lowest token budget, but this saving costs roughly four percentage points of accuracy.  

By jointly optimising for both objectives, \textsc{Premise} preserves high accuracy while substantially reducing token consumption, demonstrating the necessity of a balanced objective during prompt optimisation.
\section{Conclusion}\label{sec:conclusion}

We have presented \textbf{PREMISE}, a prompt-only framework that improves the efficiency of mathematical reasoning in Large Reasoning Models (LRMs) without touching model weights.  By coupling trace-level diagnostics for \emph{overthinking} and \emph{underthinking} with a multi-objective natural-language optimization scheme, PREMISE steers generation toward concise yet accurate solution paths.

Across GSM8K, SVAMP, and MATH-500, PREMISE matches or surpasses standard Chain-of-Thought prompting in answer accuracy while \emph{reducing reasoning tokens by as much as 87.5\% and cutting monetary cost by 69–82\%}.  These savings hold both in single-pass settings and in multi-agent systems, demonstrating that prompt-level control alone can yield substantial gains when the interface to the model is restricted to black-box API calls.

The study also reveals limitations.  When no explicit reasoning channel is exposed—illustrated by GPT-based models—the current template can lengthen the visible trace and raise cost.  Similarly, on the proof-heavy MATH-500 set with Gemini, an overly aggressive compression ratio leads to missed intermediate justifications and accuracy loss.  These cases highlight the need for \emph{adaptive compression} that aligns the token budget with task difficulty and the interface features of a given model.

Future work will extend the diagnostics to symbolic or multi-modal reasoning tasks.  We believe that such directions will further reduce inference cost while preserving the transparency and reliability expected from step-by-step reasoning.

\newpage
\clearpage
\bibliography{papers.bib} 

\onecolumn
\appendix
\section{Appendix} 

\renewcommand\thefigure{\thesection.\arabic{figure}}
\setcounter{figure}{0}

\begin{figure}[htbp]
  \centering
  \includegraphics[width=0.75\linewidth]{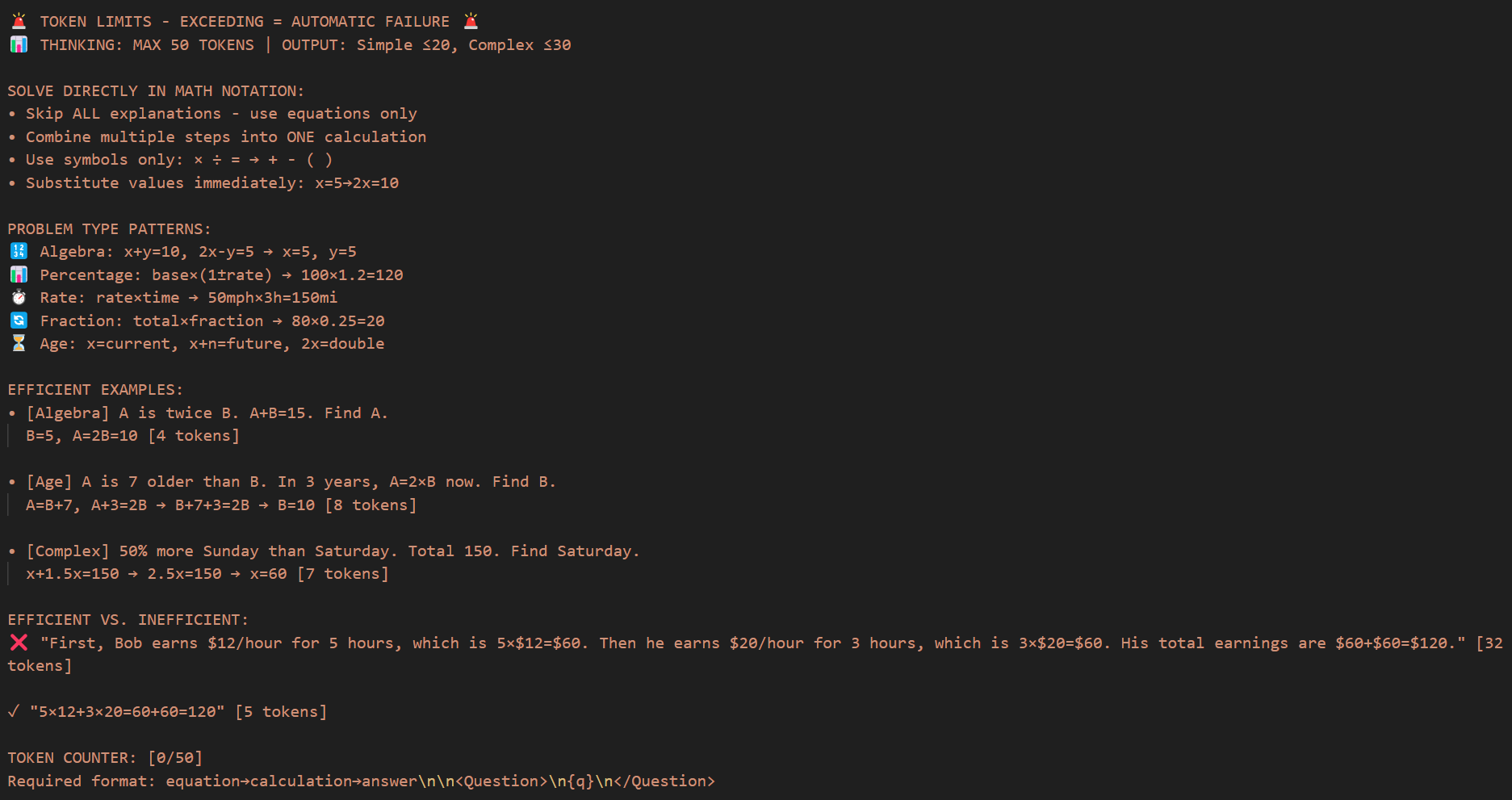}
  \caption{PREMISE Generated Efficient Reasoning Prompt}
  \label{fig:efficient-reasoning-prompt}
\end{figure}
\end{document}